\documentclass[runningheads]{llncs}
\usepackage[T1]{fontenc}
\usepackage{amsmath}
\usepackage{amsfonts}
\usepackage{amssymb}
\usepackage{mathtools}
\usepackage{graphicx}
\usepackage{algorithm}
\usepackage{algpseudocode}%

\begin{document}
\title{Neural Attention Forests: Transformer-Based Forest Improvement}
\author{Andrei V. Konstantinov \and 
Lev V. Utkin \and \\
Alexey A. Lukashin \and 
Vladimir A. Muliukha}
\authorrunning{A.V. Konstantinov et al.}
\institute{Peter the Great St.Petersburg Polytechnic University
\\St.Petersburg, Russia\\
\email{andrue.konst@gmail.com, lev.utkin@gmail.com, alexey.lukashin@spbstu.ru,vladimir.muliukha@spbstu.ru}}
\maketitle            
\begin{abstract}
A new approach called NAF (the Neural Attention Forest) for solving regression and classification tasks under tabular training data is proposed. The main idea behind the proposed NAF model is to introduce the attention mechanism into the random forest by assigning attention weights calculated by neural networks of a specific form to data in leaves of decision trees and to the random forest itself in the framework of the Nadaraya-Watson kernel regression. In contrast to the available models like the attention-based random forest, the attention weights and the Nadaraya-Watson regression are represented in the form of neural networks whose weights can be regarded as trainable parameters. The first part of neural networks with shared weights is trained for all trees and computes attention weights of data in leaves. The second part aggregates outputs of the tree networks and aims to minimize the difference between the random forest prediction and the truth target value from a training set. The neural network is trained in an end-to-end manner.  The combination of the random forest and neural networks implementing the attention mechanism forms a transformer for enhancing the forest predictions.  Numerical experiments with real datasets illustrate the proposed method. The code implementing the approach is publicly available.

\end{abstract}
%


\section{Introduction}

The attention mechanism can be viewed as an approach for the crucial
improvement of neural networks in recent years. It is based on weighing
features or examples in accordance with their importance to enhance the
machine learning model accuracy. Successful applications of the attention
mechanism to many important tasks, including natural language processing,
computer vision, etc., motivated a large interest to developing various
attention-based models
\cite{Brauwers-Frasincar-21,Chaudhari-etal-2021,Correia-Colombini-22,Niu-Zhong-Yu-21,Tay-etal-22}%
. Most attention-based models are implemented as components of neural networks
such that the attention weights are learned within the neural architectures.
In order to avoid several problems inherent in neural networks, such as
overfitting, lack of sufficient data, etc. and to use the attention mechanism
without neural networks, an attention-based random forests (ABRF) and some its
extensions have been proposed in
\cite{Konstantinov-Utkin-22d,Utkin-Konstantinov-22}. Several ideas behind the
ABRF were presented. The first one is to use random forests (RFs)
\cite{Breiman-2001} as an ensemble-based model which effectively deals with
tabular data. The problem is that neural networks successfully process various
types of data, including images, text data, graphs. However, they are inferior
to many simple models, for example, RFs, gradient boosting machines, when they
are dealing with tabular data. The second idea behind the ABRF is to apply the
well-known Nadaraya-Watson (N-W) kernel regression
\cite{Nadaraya-1964,Watson-1964} which learns a non-linear function by using a
weighted average of the data where weights are nothing else but the attention
weights \cite{Chaudhari-etal-2019,Zhang2021dive}. The third idea is to assign
trainable weights to decision trees in the RF depending on examples, which
fall into leaves of trees, and their features. The weights are trained by
solving a simple quadratic optimization problem which is constructed as a
standard loss function to minimize the classification or regression errors.
The ABRF have demonstrated outperforming results on many real datasets. In
spite of the ABRF efficiency, these models use the standard Gaussian kernels
with trainable parameters to train the attention weights. Moreover, the set of
trainable attention parameters is rather narrow and does not allow us to
significantly enhance the predicted accuracy of the model.

In order to overcome the above difficulties, we propose to combine the RF and
the neural network of a specific architecture. Actually, the neural network
implements a set of attention operations and consists of two parts. The first
part of neural networks with shared weights is trained for all trees and
computes attention weights of data in leaves. The second part aggregates
outputs of the \textquotedblleft tree\textquotedblright \ networks and aims to
minimize the difference between the random forest prediction and the truth
target value from a training set. These parts are jointly trained in an
end-to-end manner. The combination of the random forest and neural networks
implementing the attention mechanism forms a transformer for enhancing the
forest predictions.

Our contributions can be summarized as follows. A new Neural Attention Forest
architecture called as \emph{NAF} is proposed to overcome difficulties of RFs
and neural networks when we deal with tabular data. The attention mechanism is
implemented by two different parts of the neural network. The first part
computes trainable attention weights depending on trees and examples. The
second part of the network aggregates weighted outputs of trees. Numerical
experiments with real datasets are performed for studying NAF. They
demonstrate outperforming results of NAF in comparison with RFs. The code of
proposed algorithms can be found at https://github.com/andruekonst/NAF.

The paper is organized as follows. A brief introduction to the attention
mechanism as the N-W kernel regression and the attention-based random forest
is given in Section 2. A general architecture of NAF is considered in Section
3. The transformer implementation of NAF is provided in Section 4. Numerical
experiments with real data illustrating properties of NAF are provided in
Section 5. Concluding remarks can be found in Section 6.

\section{Preliminaries}

\subsection{Nadaraya-Watson regression and attention}

Suppose that a dataset $\mathcal{D}$ is represented by $n$ examples
$(\mathbf{x}_{1},y_{1}),...,(\mathbf{x}_{n},y_{n})$, where $\mathbf{x}%
_{i}=(x_{i1},...,x_{id})\in \mathbb{R}^{d}$ is a feature vector; $y_{i}%
\in \mathbb{R}$ is a regression output. The regression task is to construct a
regressor $f:\mathbb{R}^{d}\rightarrow \mathbb{R}$ which can predict the output
value $\widehat{y}$ of a new observation $\mathbf{x}$, using the dataset. The
N-W kernel regression model \cite{Nadaraya-1964,Watson-1964} is one of the
methods to estimate the function $f$ by applying the weighted averaging as
follows:
\begin{equation}
\widehat{y}=\sum_{i=1}^{n}\alpha(\mathbf{x},\mathbf{x}_{i})y_{i},
\end{equation}
where weight $\alpha(\mathbf{x},\mathbf{x}_{i})$ conforms with relevance of
the feature vector $\mathbf{x}_{i}$ to the vector $\mathbf{x}$, i.e., the
closer $\mathbf{x}_{i}$ to $\mathbf{x}$, the greater the weight $\alpha
(\mathbf{x},\mathbf{x}_{i})$ assigned to $y_{i}$.

Weights are calculated by using a normalized kernel $K(\mathbf{x}%
,\mathbf{x}_{i})$ as:
\begin{equation}
\alpha(\mathbf{x},\mathbf{x}_{i})=\frac{K(\mathbf{x},\mathbf{x}_{i})}%
{\sum_{j=1}^{n}K(\mathbf{x},\mathbf{x}_{j})}.
\end{equation}

According to \cite{Bahdanau-etal-14}, vector $\mathbf{x}$, vectors
$\mathbf{x}_{i}$, outputs $y_{i}$, and weight $\alpha(\mathbf{x}%
,\mathbf{x}_{i})$ are called as the \textit{query}, \textit{keys},
\textit{values,} and the \textit{attention weight}, respectively. Weights
$\alpha(\mathbf{x},\mathbf{x}_{i})$ can be extended by incorporating trainable
parameters. Several types of attention weights have been proposed in
literature. The most important examples are the additive attention
\cite{Bahdanau-etal-14}, multiplicative or dot-product attention
\cite{Luong-etal-2015,Vaswani-etal-17}.

\subsection{Attention-based random forest}

A powerful machine learning model handling tabular data is the RF which is
represented as an ensemble of $T$ trees such that each tree is trained on a
subset of examples randomly selected from the training set. A prediction
$\widehat{y}$ for a new example $\mathbf{x}$ in the RF is determined by
averaging predictions $\widehat{y}_{1},...,\widehat{y}_{T}$ obtained for all
trees in the RF. Let us denote an index set of examples, which jointly with
$\mathbf{x}$ fall into the same leaf in the $k$-th tree, as $\mathcal{J}%
_{k}(\mathbf{x})$. Then the corresponding leaf can be characterized by the
vector $\mathbf{A}_{k}(\mathbf{x)}$ which is the mean of all $\mathbf{x}_{j}$
from $\mathcal{D}$ such that $j\in \mathcal{J}_{k}(\mathbf{x})$, and the vector
$B_{k}(\mathbf{x)}$ which is a mean of all $y_{j}$ from $\mathcal{D}$ by
$j\in \mathcal{J}_{k}(\mathbf{x})$. According to ABRF
\cite{Utkin-Konstantinov-22}, the N-W regression can be rewritten in terms of
the RF as:%
\begin{equation}
\widehat{y}=\sum_{k=1}^{T}\alpha \left(  \mathbf{x},\mathbf{A}_{k}%
(\mathbf{x)},\theta \right)  \cdot B_{k}(\mathbf{x)},\label{RF_Att_47}%
\end{equation}
where $\theta$ is a vector of training attention parameters.

It follows from definitions of the attention mechanism that $B_{k}%
(\mathbf{x)}$, $\mathbf{A}_{k}(\mathbf{x)}$, and $\mathbf{x}$ are the
\textit{value}, the \textit{key}, and the \textit{query}, respectively. If to
return to the original RF, then all its trees have the same weights
$\alpha \left(  \mathbf{x},\mathbf{A}_{k}(\mathbf{x)},\theta \right)  =1/T$.
Parameters $\theta$ are trained by minimizing the expected loss function $L$
over a set $\Theta$ of parameters as follows:
\begin{equation}
\theta_{opt}=\arg \min_{\theta \mathbf{\in}\Theta}~\sum_{s=1}^{n}L\left(
\widehat{y}_{s},y_{s},\theta \right)  ,\label{ABRF-10}%
\end{equation}
where $\widehat{y}_{s}$, $y_{s}$ are the truth target value and the predicted
output of the $s$-th input example, respectively.

The original ABRF uses the Gaussian kernel for computing the attention weights
and the Huber's $\epsilon$-contamination model \cite{Huber81} to reduce the
problem (\ref{ABRF-10}) to the standard quadratic optimization problem with
linear constraints. However, the obtained approach cannot cover a wide range
of possible kernels and models for computing $\mathbf{A}_{k}(\mathbf{x)}$,
$B_{k}(\mathbf{x)}$ and other parameters. Moreover, the set of trainable
parameters $\theta$ in the ABRF is also restrictive. Therefore, we propose a
general approach which uses neural networks to compute the attention weights
simultaneously with $\mathbf{A}_{k}(\mathbf{x)}$ and $B_{k}(\mathbf{x)}$ such
that the neural network weights are trainable attention parameters $\theta$.

\section{The neural attention forest architecture}

\begin{figure}
[ptb]
\begin{center}
\includegraphics[
height=3.2699in,
width=3.7764in
]%
{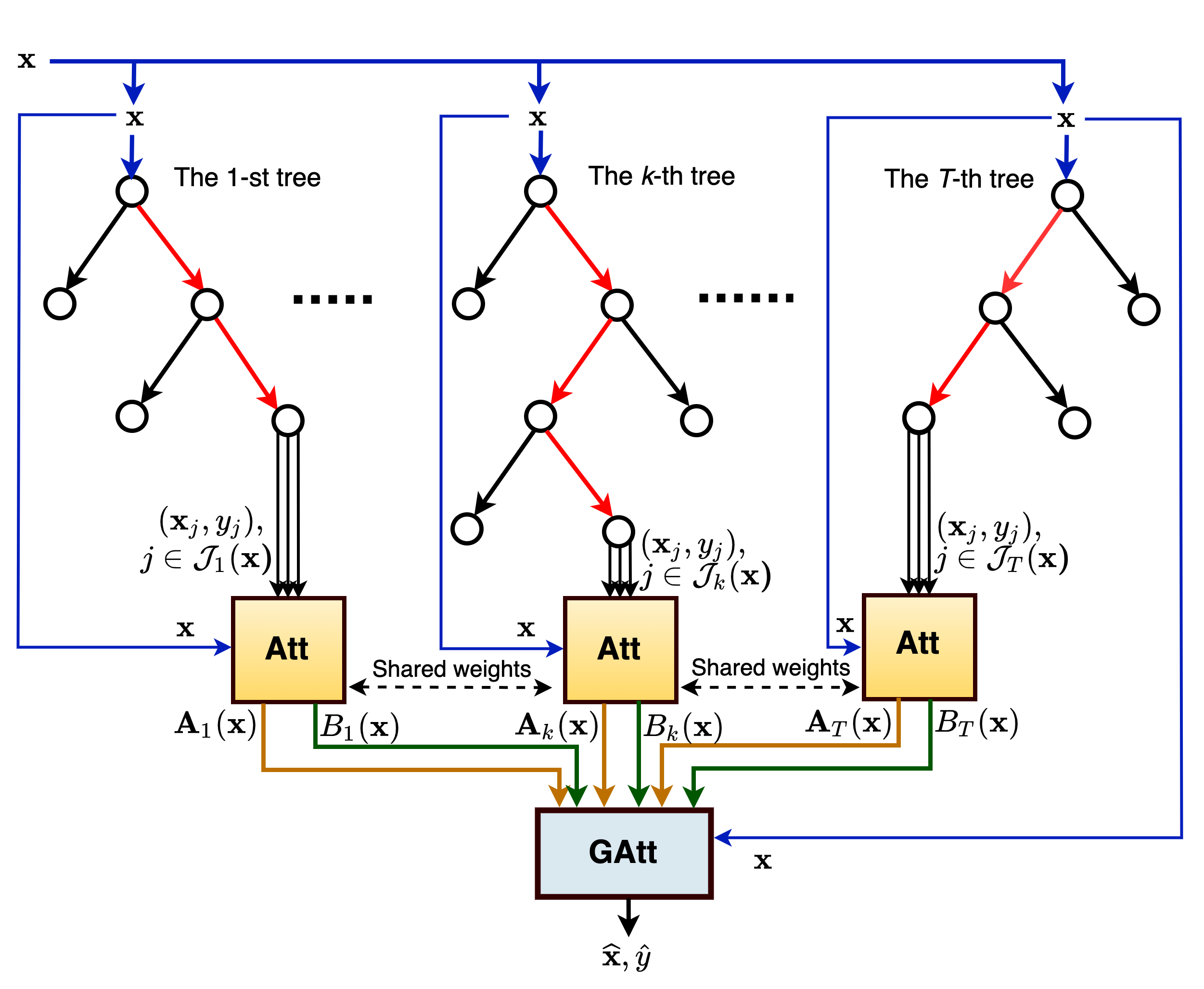}%
\caption{The neural attention forest architecture}%
\label{f:Gen_Scheme}%
\end{center}
\end{figure}

A general NAF architecture is shown in Fig.\ref{f:Gen_Scheme}. Suppose that we
have built a RF consisting of $T$ decision trees on the training set
$\mathcal{D}$. A feature vector $\mathbf{x}$ is fed to each tree, and the path
of $\mathbf{x}$ to a leaf is depicted by the red line. Several feature vectors
$\mathbf{x}_{j}$ as well as the corresponding output values $y_{j}$ fall into
the same leaf with $j\in \mathcal{J}_{k}(\mathbf{x})$. The first part of the
neural network (Att in Fig.\ref{f:Gen_Scheme}) implementing the attention
mechanism is represented by $T$ networks with shared weights $\theta$
(parameters of the networks). It can be seen from Fig.\ref{f:Gen_Scheme} that
the $k$-th neural network implements the attention operation and computes
vector $\mathbf{A}_{k}(\mathbf{x)}$ and value $B_{k}(\mathbf{x)}$ in
accordance with the N-W regression as follows:
\begin{equation}
\mathbf{A}_{k}(\mathbf{x)}=\sum_{j\in \mathcal{J}_{k}(\mathbf{x})}\alpha \left(
\mathbf{x},\mathbf{x}_{j},\theta \right)  \mathbf{x}_{j},\label{RF_Att_L_20}%
\end{equation}%
\begin{equation}
B_{k}(\mathbf{x)}=\sum_{j\in \mathcal{J}_{k}(\mathbf{x})}\alpha \left(
\mathbf{x},\mathbf{x}_{j},\theta \right)  y_{j}.\label{RF_Att_L_21}%
\end{equation}
where $\alpha \left(  \mathbf{x},\mathbf{x}_{j},\theta \right)  $ is the
attention weight having trainable parameters $\theta$ (parameters of the
networks) and is defined by the following scaled dot-product\textbf{\ }score
function \cite{Luong-etal-2015}: $\alpha(\mathbf{x},\mathbf{x}_{j}%
)=\mathbf{x}^{T}\mathbf{x}_{j}\mathbf{/}\sqrt{d}$.

In sum, we get keys $\mathbf{A}_{k}(\mathbf{x)}$ and values $B_{k}%
(\mathbf{x)}$ for all trees. The second part of the neural network is the
global attention (GAtt in Fig.\ref{f:Gen_Scheme}) which aggregates all keys
$\mathbf{A}_{k}(\mathbf{x)}$ and values $B_{k}(\mathbf{x)}$ in accordance with
the N-W regression as follows:%
\begin{equation}
\widehat{\mathbf{x}}=\sum_{k=1}^{T}\beta \left(  \mathbf{x},\mathbf{A}%
_{k}(\mathbf{x)},\psi \right)  \cdot \mathbf{A}_{k}(\mathbf{x)},\label{GAtt_1}%
\end{equation}%
\begin{equation}
\widehat{y}=\sum_{k=1}^{T}\beta \left(  \mathbf{x},\mathbf{A}_{k}%
(\mathbf{x)},\psi \right)  \cdot B_{k}(\mathbf{x)},\label{GAtt_2}%
\end{equation}
where $\beta$ is the attention weight with parameters $\psi$ (parameters of
the second part of the network).

Note that the accurately trained NAF predicts vector $\widehat{\mathbf{x}}$
which should be close to $\mathbf{x}$. By comparing $\widehat{\mathbf{x}}$ and
$\mathbf{x}$, we can judge the quality of training the network and the RF. NAF
is learned in an end-to-end manner by using the loss function%
\begin{equation}
(\theta,\psi)_{opt}=\arg \min_{\theta,\psi}~\sum_{s=1}^{n}\left(  \widehat
{y}_{s}-y_{s}\right)  ^{2}.
\end{equation}
%

\section{The neural attention forest as a transformer}

Let us represent the set of examples $(\mathbf{x}_{j},y_{j})$, $j\in
\mathcal{J}_{k}(\mathbf{x})$, which fall jointly with the feature vector
$\mathbf{x}$ in the same $p$-th leaf of the $k$-th tree, as a sentence
$\mathbf{s}$ consisting of words denoted as $w_{i}^{(k,p)}$, $i\in
\mathcal{J}_{k}(\mathbf{x})$, i.e.,
\begin{equation}
\mathbf{s}_{k,p}=\left \{  w_{i_{1}}^{(k,p)},...,w_{i_{r}}^{(k,p)}\right \}  .
\end{equation}

\begin{figure}
[ptb]
\begin{center}
\includegraphics[
height=2.8954in,
width=2.7683in
]%
{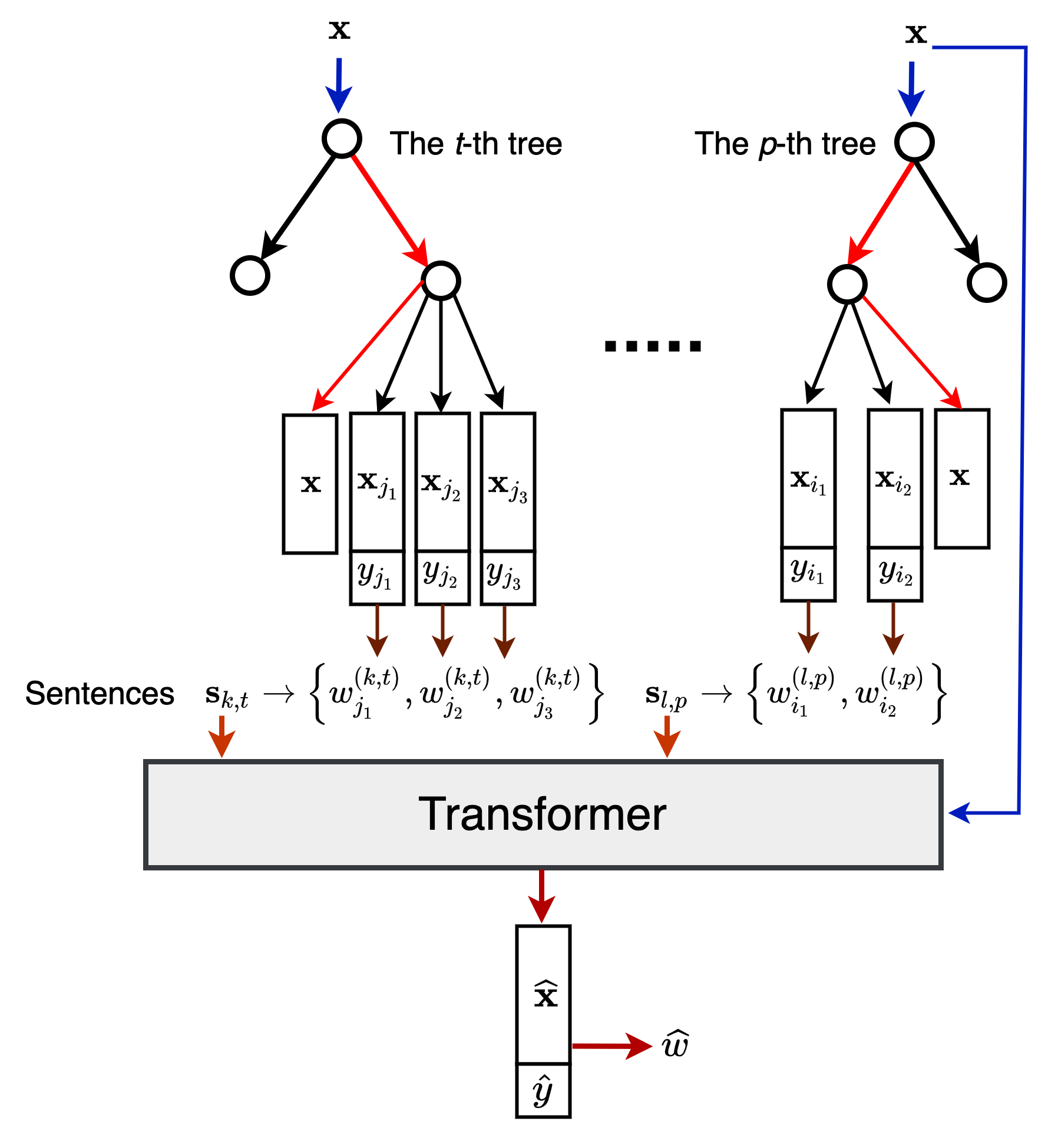}%
\caption{A scheme how sentences for the transformer are produced by trees }%
\label{f:Transformer}%
\end{center}
\end{figure}

Here $r$ is the number of elements in $\mathcal{J}_{k}(\mathbf{x})$; $k$ is
the number of a tree; $p$ is the number of the leaf; $i$ is the example index
(the word index in the $p$-th sentence), which falls into the $p$-th leaf.
Sentences $\mathbf{s}_{k,t}$ and $\mathbf{s}_{l,p}$ produced by the $t$-th and
the $p$-th trees and consisting of $3$ and $2$ words, respectively, as an
illustrative example are shown in Fig.\ref{f:Transformer}.

The above implies that every leaf produces a sentence $\mathbf{s}_{k,p}$. In
sum, we have a set $\mathcal{S}$ of sentences $s$ produced by all leaves.
Query $\mathbf{x}$ can also be regarded as a word $\overline{w}=(\mathbf{x})$
which consists of $\mathbf{x}$ and does not contain $y$ that has to be estimated.

Let us construct a model
\begin{equation}
\mathcal{M}_{\theta}\left(  \mathbf{s}_{k,p},\overline{w}\right)
=\mathcal{M}_{\theta}\left(  w_{i_{1}}^{(k,p)},...,w_{i_{r}}^{(k,p)}%
;\overline{w}\right)  .
\end{equation}

The above notation means that $\overline{w}$ is the query, $w_{i_{1}}%
^{(k,p)},...,w_{i_{r}}^{(k,p)}$ are pairs of keys $\mathbf{x}_{j}$ and values
$y_{j}$, $j\in \mathcal{J}_{k}(\mathbf{x})$, in the attention model. In other
words, we compute $\mathbf{A}_{k}(\mathbf{x)}$ and $B_{k}(\mathbf{x)}$ by
using $\mathcal{M}_{\theta}\left(  \mathbf{s}_{k,p};\overline{w}\right)  $ in
accordance with (\ref{RF_Att_L_20}) and (\ref{RF_Att_L_21}).

Denote the set of all words defined by the training set $\mathcal{D}$ as
$\mathcal{W}$. If we consider only the level of the first neural networks,
then an optimization problem for learning parameters $\theta$ of
$\mathcal{M}_{\theta}$ can be written as
\begin{equation}
\sum_{w\in \mathcal{W}}\sum_{\mathbf{s}\in \mathcal{S}:w\in \mathbf{s}}\left \Vert
\mathcal{M}_{\theta}\left(  \mathbf{s}\backslash w;\overline{w}\right)
-w\right \Vert ^{2}\rightarrow \min_{\theta}.
\end{equation}

However, the second part of the neural network should be taken into account.
Therefore, we have to aggregate the models $\mathcal{M}_{\theta}$ obtained for
all trees. Let us return to the attention operations (\ref{GAtt_1}) and
(\ref{GAtt_1}). Then the loss function can be rewritten as
\begin{equation}
\sum_{w\in \mathcal{W}}\left \Vert Q\left(  \widehat{w}-w\right)  \right \Vert
^{2}=\sum_{w\in \mathcal{W}}\left \Vert Q\left(  \sum_{t=1}^{T}\beta \left(
\widetilde{w}_{t}\right)  \cdot \widetilde{w}_{t}-w\right)  \right \Vert ^{2},
\end{equation}
where $\widetilde{w}_{t}$ is defined as
\[
\widetilde{w}_{t}=\mathcal{M}_{\theta}\left(  \mathbf{s}^{(t)}(w)\backslash
w;\overline{w}\right)  ,
\]
$\mathbf{s}^{(t)}(\overline{w})$ is a sentence produced by the $t$-th tree,
which contains the word $w$ (there exists exactly one such sentence); $Q$ is a
$(d+1)\times(d+1)$ diagonal matrix whose diagonal is $(\lambda_{1},\lambda
_{2},...,\lambda_{d},1)$; $0\leq \lambda_{i}\leq1$ is the hyperparameter which
determines the weight of the $i$-th feature in the loss function, the last $1$
in the diagonal of matrix $Q$ indicates that $y$ from the word $w$ and
$\widehat{y}$ from $\widehat{w}$ have to be used.%

\section{Numerical experiments}

In order to study NAF, we build two types of the forests. The first one is the
original RF. The second forest is the Extremely Randomized Trees (ERT)
proposed by Geurts et al. \cite{Geurts-etal-06}. In contrast to RFs, the ERT
algorithm at each node chooses a split point randomly for each feature and
then selects the best split among these features.

In all experiments, RFs as well as ERTs consist of $100$ trees. A 3-fold
cross-validation on the training set consisting of $n_{\text{tr}}=4n/5$
examples with $100$ repetitions is performed. The testing set for computing
the accuracy measures consists of $n_{\text{test}}=n/5$ examples. In order to
get desirable estimates of vectors $\mathbf{A}_{k}(\mathbf{x})$ and values
$B_{k}(\mathbf{x})$, all trees in experiments are trained such that at least
$10$ examples fall into every leaf of a tree. The coefficient of determination
denoted as $R^{2}$ is used for the regression evaluation. The greater the
value of the coefficient of determination, the better results we get.

NAF is investigated by applying datasets which are taken from open sources.
The dataset Diabetes is downloaded from the R Packages; datasets Friedman 1,
2, 3, Regression and Sparse are taken from package \textquotedblleft
Scikit-Learn\textquotedblright; datasets Boston Housing (Boston) and Yacht
Hydrodynamics (Yacht) can be found in the UCI Machine Learning Repository
\cite{Dua:2019}.

Values of the measure $R^{2}$ for several models, including RF, ERT, NAF with
one layer having 16 units and linear activations added by the softmax
operation (NAF-1), and NAF with three layers having 16 units in two layers
with the tanh activation function and one linear layer (NAF-3) are shown in
Table \ref{t:Compar_Models_1} which also contain a brief information about the
datasets (the number of features $m$ and the number of examples $n$). Results
presented in Table \ref{t:Compar_Models_1} are given for two cases when the RF
and the ERT are used. The best results separately for NAFs based on the RF and
the ERT are shown in bold. It should be pointed out that NAF demonstrates
outperforming results for most dataset. Moreover, one can see from Table
\ref{t:Compar_Models_1} that NAF based on the ERT provides better results in
comparison the NAF based on the RF. A very important comment with respect to
the obtained results is that we have used the simplest neural network having
one layer (NAF-1). Even this simple network has ensured the outperforming results.%

\begin{table}[tbp] \centering
\caption{Values of $R^2$ for comparison of models based on the RF and the
ERT}%
\begin{tabular}
[c]{ccccccccc}\hline
&  &  & \multicolumn{3}{c}{RF} & \multicolumn{3}{c}{ERT}\\ \hline
Data set & $m$ & $n$ & Original & NAF-1 & NAF-3 & Original & NAF-1 &
NAF-3\\ \hline
Diabetes & $10$ & $442$ & $0.417$ & $0.382$ & $\mathbf{0.418}$ & $0.438$ &
$0.396$ & $\mathbf{0.442}$\\
Friedman 1 & $10$ & $100$ & $0.459$ & $\mathbf{0.506}$ & $0.487$ & $0.471$ &
$0.514$ & $\mathbf{0.530}$\\
Friedman 2 & $4$ & $100$ & $0.840$ & $\mathbf{0.903}$ & $0.892$ & $0.813$ &
$\mathbf{0.943}$ & $0.921$\\
Friedman 3 & $4$ & $100$ & $0.625$ & $\mathbf{0.720}$ & $0.635$ & $0.570$ &
$\mathbf{0.608}$ & $0.578$\\
Boston & $13$ & $506$ & $0.814$ & $\mathbf{0.846}$ & $0.832$ & $0.831$ &
$\mathbf{0.857}$ & $0.855$\\
Yacht & $6$ & $308$ & $0.981$ & $0.976$ & $\mathbf{0.985}$ & $0.988$ &
$\mathbf{0.995}$ & $0.992$\\
Regression & $100$ & $100$ & $\mathbf{0.380}$ & $0.291$ & $0.346$ & $0.402$ &
$0.333$ & $\mathbf{0.406}$\\
Sparse & $10$ & $100$ & $0.470$ & $\mathbf{0.671}$ & $0.613$ & $0.452$ &
$\mathbf{0.701}$ & $0.665$\\ \hline
\end{tabular}
\label{t:Compar_Models_1}%
\end{table}%

\begin{figure}
[ptb]
\begin{center}
\includegraphics[
height=1.6086in,
width=4.7298in
]%
{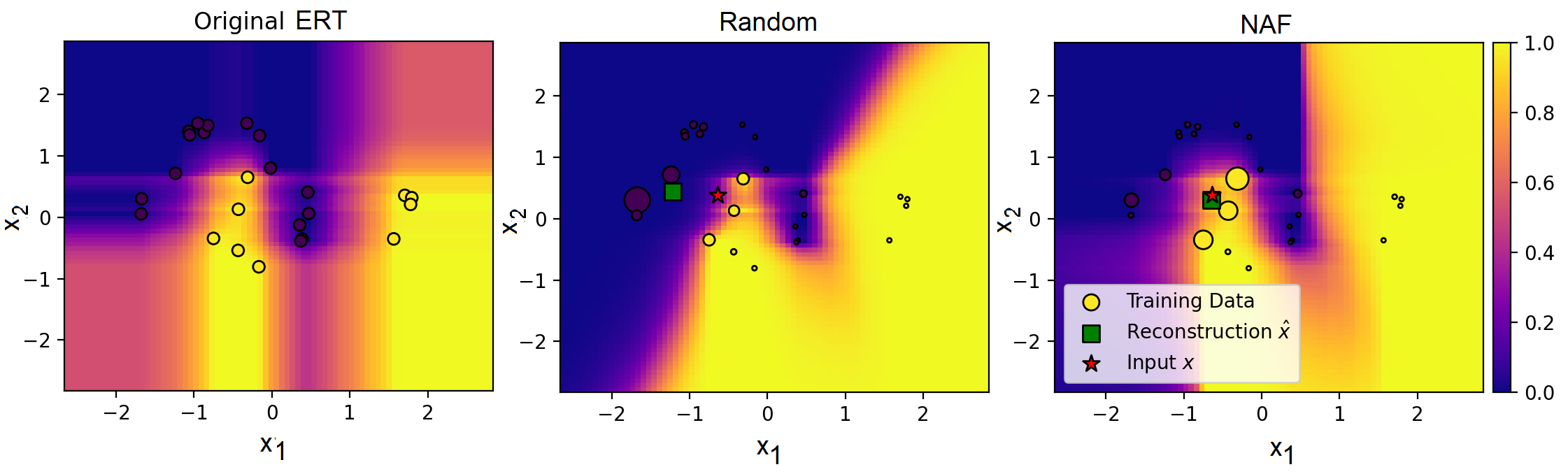}%
\caption{Illustration of the example-based explanation on the
\textquotedblleft two moons\textquotedblright \ dataset}%
\label{f:xai_moons}%
\end{center}
\end{figure}

One of the advantages of NAF is its simple example-based explanation which
selects instances of the dataset to explain the prediction. The idea behind
the explanation is to use the property of NAF to predict (or reconstruct)
$\widehat{\mathbf{x}}$ which is close to input vector $\mathbf{x}$. If the
reconstructed vector $\widehat{\mathbf{x}}$ is actually close to the input
vector $\mathbf{x}$, then the nearest neighbors for the reconstructed vector
can be regarded as examples explaining a prediction. For the explanation
experiments, we build 500 trees of the ERT on the well-known \textquotedblleft
two moons\textquotedblright \ dataset. Moreover, we use a condition that at
least $1$ example falls into every leaf of a tree. $25$ points are taken for
training. The measure $R^{2}$ is $0.781$ for the ERT, and is $0.936$ for NAF.
Fig.\ref{f:xai_moons} illustrates results provided by three models: the
original ERT (the left picture), the random NAF with random weights of the
neural network (the middle picture), and NAF-1 (the right picture). First, it
can be seen from the middle picture that the reconstructed vector depicted by
the small square is far from the input vector depicted by the small star
because the network with random weights provides incorrect results. Therefore,
the neighbors of $\widehat{\mathbf{x}}$ cannot be viewed as examples for
explanation. However, it can be seen from the right picture in
Fig.\ref{f:xai_moons} that NAF provides the reconstructed vector which is
close to the input vector. Therefore, its nearest neighbors can be used for
the example-based explanation. The nearest neighbors are also defined by their
weight (the small circle size) calculated as product of two attention weights
$\alpha \left(  \mathbf{x},\mathbf{A}_{k}(\mathbf{x)},\theta \right)  $ and
$\beta \left(  \mathbf{x},\mathbf{A}_{k}(\mathbf{x)},\psi \right)  $.%

\section{Concluding remarks}

A new approach to combining RFs and attention-based neural networks for
solving regression and classification tasks under tabular training data has
been proposed. It has been shown that the proposed model can be represented as
a transformer-based model which improves the RF. Numerical experiments with
real datasets have demonstrated that NAF provides outperforming results even
when simple one-layer neural networks are used to implement the attention operations.

NAF opens a door for developing a new class of neural attention forest models
and a new class of transformers. One of the interesting directions for further
research is to consider the attention when different RFs are built with the
same data. Moreover, this idea leads to implementation of the multi-head
attention and the cross-attention depending on a scheme of training the
corresponding RFs. Another direction for research is to consider the NAF as
an autoencoder which transforms the input vector $\mathbf{x}$ into its
estimates $\widehat{\mathbf{x}}$. By using this representation, we can
construct a new anomaly detection procedure under condition of small data. It
should be noted that NAF can also be a basis for developing a weakly
supervised model. First, we learn trees by using a small labeled dataset. Then
we learn the neural network on unlabeled data minimizing the difference
between reconstructed vector $\widehat{\mathbf{x}}$ and the input vector
$\mathbf{x}$ as a loss function. Then the unlabeled examples can be regarded
as new tested examples, and their labels can be estimated by NAF. This is a
perspective direction for research.

The above directions is a part of many improvements and modifications of NAF,
which could significantly extend the class of the neural attention forest models.

\bibliographystyle{splncs04}
\bibliography{Attention,Boosting,Classif_bib,Deep_Forest,Imprbib,MYBIB,MYUSE}

\begin{thebibliography}{10}
\providecommand{\url}[1]{\texttt{#1}}
\providecommand{\urlprefix}{URL }
\providecommand{\doi}[1]{https://doi.org/#1}

\bibitem{Bahdanau-etal-14}
Bahdanau, D., Cho, K., Bengio, Y.: Neural machine translation by jointly
  learning to align and translate (Sep 2014), arXiv:1409.0473

\bibitem{Brauwers-Frasincar-21}
Brauwers, G., Frasincar, F.: A general survey on attention mechanisms in deep
  learning. IEEE Transactions on Knowledge and Data Engineering  (2021)

\bibitem{Breiman-2001}
Breiman, L.: Random forests. Machine learning  \textbf{45}(1),  5--32 (2001)

\bibitem{Chaudhari-etal-2019}
Chaudhari, S., Mithal, V., Polatkan, G., Ramanath, R.: An attentive survey of
  attention models (Apr 2019), arXiv:1904.02874

\bibitem{Chaudhari-etal-2021}
Chaudhari, S., Mithal, V., Polatkan, G., Ramanath, R.: An attentive survey of
  attention models. ACM Transactions on Intelligent Systems and Technology
  \textbf{12}(5),  1--32 (2021), article 53

\bibitem{Correia-Colombini-22}
Correia, A., Colombini, E.: Attention, please! {A} survey of neural attention
  models in deep learning. Artificial Intelligence Review  \textbf{55}(8),
  6037--6124 (2022)

\bibitem{Dua:2019}
Dua, D., Graff, C.: {UCI} machine learning repository (2017),
  \url{http://archive.ics.uci.edu/ml}

\bibitem{Geurts-etal-06}
Geurts, P., Ernst, D., Wehenkel, L.: Extremely randomized trees. Machine
  learning  \textbf{63},  3--42 (2006)

\bibitem{Huber81}
Huber, P.: Robust Statistics. Wiley, New York (1981)

\bibitem{Konstantinov-Utkin-22d}
Konstantinov, A., Utkin, L., Kirpichenko, S.: {AGB}oost: Attention-based
  modification of gradient boosting machine. In: 31st Conference of Open
  Innovations Association (FRUCT). pp. 96--101. IEEE (2022).
  \doi{10.23919/FRUCT54823.2022.9770928}

\bibitem{Luong-etal-2015}
Luong, T., Pham, H., Manning, C.: Effective approaches to attention-based
  neural machine translation. In: Proceedings of the 2015 Conference on
  Empirical Methods in Natural Language Processing. pp. 1412--1421. The
  Association for Computational Linguistics (2015)

\bibitem{Nadaraya-1964}
Nadaraya, E.: On estimating regression. Theory of Probability \& Its
  Applications  \textbf{9}(1),  141--142 (1964)

\bibitem{Niu-Zhong-Yu-21}
Niu, Z., Zhong, G., Yu, H.: A review on the attention mechanism of deep
  learning. Neurocomputing  \textbf{452},  48--62 (2021)

\bibitem{Tay-etal-22}
Tay, Y., Dehghani, M., Bahri, D., Metzler, D.: Efficient transformers: A
  survey. ACM Computing Surveys  \textbf{55}(6),  1--28 (2022)

\bibitem{Utkin-Konstantinov-22}
Utkin, L., Konstantinov, A.: Attention-based random forest and contamination
  model. Neural Networks  \textbf{154},  346--359 (2022)

\bibitem{Vaswani-etal-17}
Vaswani, A., Shazeer, N., Parmar, N., Uszkoreit, J., Jones, L., Gomez, A.,
  Kaiser, L., Polosukhin, I.: Attention is all you need. In: Advances in Neural
  Information Processing Systems. pp. 5998--6008 (2017)

\bibitem{Watson-1964}
Watson, G.: Smooth regression analysis. Sankhya: The Indian Journal of
  Statistics, Series A pp. 359--372 (1964)

\bibitem{Zhang2021dive}
Zhang, A., Lipton, Z., Li, M., Smola, A.: Dive into deep learning.
  arXiv:2106.11342  (Jun 2021)

\end{thebibliography}

\end{document}